\documentclass[conference]{IEEEtran}

\usepackage[utf8]{inputenc}       
\usepackage{cite}                 
\usepackage{amsmath}              
\usepackage{graphicx}             
\usepackage{url}                  
\usepackage[colorlinks=true,
            allcolors=black]{hyperref}  
\usepackage{caption}
\usepackage{subcaption}
\usepackage{float}
\usepackage{booktabs}
\usepackage{tabularx}
\usepackage{makecell}

\title{Adaptive Joint Compression and Synchronisation in Federated Split Learning for IoT Rainfall Prediction}

\author{

\IEEEauthorblockN{
Wenjie Ding,
Yi Sin Lin,
Jiale Liu,
Baoyi Liu,
Guanghua Liu,
Zhuolu Li,\\
Suleiman Sabo,
Chuadhry Mujeeb Ahmed,
Aydin Abadi,
Rehmat Ullah,
Rajiv Ranjan
}
\IEEEauthorblockA{
\textit{School of Computing} \\
\textit{Newcastle University} \\
Newcastle upon Tyne, United Kingdom \\
\{W.Ding7, Y.Lin69, J.Liu165, B.Liu25, G.Liu15, Z.Li174, S.M.Sabo2,\\
Mujeeb.Ahmed, Aydin.Abadi, Rehmat.Ullah, Raj.Ranjan\}@newcastle.ac.uk
}
}

\begin{document}
\maketitle

\begin{abstract}
Federated split learning (FSL) enables collaborative training across bandwidth-constrained IoT devices, but repeated activation and gradient exchange creates a communication bottleneck. Prior work optimises either activation compression or synchronisation frequency in isolation. This paper presents an FSL framework for IoT rainfall prediction that jointly regulates activation compression and the synchronisation interval $\rho$ via a latency driven scheduler on a server with per client EMA smoothing. The system is evaluated on hourly ERA5 data from 11 weather stations through a 17 scenario simulation matrix and a four scenario Raspberry Pi deployment over a real wide-area link. The simulation matrix validates scheduler switching across low, high, and mixed latency profiles, while the Pi deployment validates the high latency endpoint selected by the same policy. AUPRC varies only slightly across configurations (0.6381--0.6484 in simulation; within 0.011 on Pi), indicating that aggressive quantisation and sparser aggregation do not materially degrade predictive quality in this setting. On Pi, the selected endpoint (int8 with $\rho{=}3$) achieves an 87\% reduction in activation upload payload and a 54\% reduction in synchronisation traffic relative to the float32 baseline, while reducing runtime jitter from $\pm$688~s to $\pm$10~s.
\end{abstract}


\section{Introduction}
\label{sec:intro}

Internet of Things (IoT) sensing networks are widely used in smart city and environmental monitoring applications, generating large volumes of time-series observations such as rainfall, temperature, and humidity~\cite{GUBBI20131645,6740844} that support data-driven rainfall prediction~\cite{CHAN2023100375,10.3389/frwa.2024.1378598}. Conventional centralised machine learning, however, requires frequent transmission of raw data to cloud servers, raising privacy concerns and incurring high communication overhead in bandwidth-constrained IoT environments~\cite{pmlr-v54-mcmahan17a,2019arXiv191204977K}. Federated learning (FL) and federated split learning (FSL) have emerged as distributed paradigms that enable collaborative training without raw data sharing~\cite{pmlr-v54-mcmahan17a,thapa2022splitfedfederatedlearningmeets,vepakomma2018splitlearninghealthdistributed}, but FSL itself introduces a communication bottleneck because clients repeatedly transmit intermediate activations and receive gradients during training~\cite{vepakomma2018splitlearninghealthdistributed,thapa2022splitfedfederatedlearningmeets}. This overhead is mainly determined by two coupled factors: activation payload size and client–server synchronisation frequency. Each factor governs the trade-off between communication cost and model fidelity or convergence. Existing approaches typically address these factors independently~\cite{shiranthika2024splitfedziplearnedcompressiondata,11458714}, leaving the effectiveness of joint, adaptive control in multi-node FSL systems underexplored.

To address this gap, we propose an FSL framework for IoT rainfall prediction that explores the trade-off between communication efficiency and predictive performance, equipped with an adaptive mechanism that jointly adjusts compression and the synchronisation interval at runtime in response to latency signals. The simulation scenarios evaluate runtime switching under controlled heterogeneous conditions, while the Pi deployment tests the high latency endpoint selected by the same policy on real devices. The main contributions of this work are summarised as follows:
\begin{enumerate}
    \item A communication efficient federated split learning framework for rainfall prediction using distributed IoT weather data.
    \item A joint optimisation approach that combines activation compression with tunable synchronisation intervals to reduce activation upload payload and server aggregation workload.
    \item An adaptive scheduling mechanism that adjusts communication parameters based on runtime latency signals, with validation on real hardware for the selected high latency endpoint.
    \item An experimental evaluation that analyses the trade-off among communication efficiency, end-to-end runtime, and prediction performance.
\end{enumerate}

\section{Related Work}
\label{sec:RelatedWork}

\subsection{Communication Efficient Federated Split Learning}
In FL, communication mainly comes from repeated exchanges of model updates between clients and the server. FedAvg reduces this overhead by performing multiple local SGD epochs before aggregation, thereby reducing the number of communication rounds~\cite{pmlr-v54-mcmahan17a}. Communication in Federated Split Learning (FSL) differs: clients transmit intermediate activations, also called smashed data, and receive their gradients during backpropagation~\cite{vepakomma2018splitlearninghealthdistributed,thapa2022splitfedfederatedlearningmeets}. Since the cost depends on the cut layer, activation shape, and batch size, FL-oriented compression methods cannot always be directly applied to FSL.

Recent studies have explored communication efficient compression for SL and SplitFed systems. SplitFedZip introduces learned rate distortion compression for features and gradients in SplitFed learning~\cite{shiranthika2024splitfedziplearnedcompressiondata}. SL-ACC applies adaptive channel wise quantisation by estimating channel importance with entropy~\cite{11458714}, while SL-FAC uses frequency decomposition and frequency based quantisation to preserve important spectral components~\cite{lin2026slfaccommunicationefficientsplitlearning}. These methods show that smashed data compression can reduce communication overhead, but they mainly focus on compression itself rather than jointly controlling compression and synchronisation frequency during runtime.

\subsection{Adaptive Scheduling and System Level Optimisation}

In addition to activation compression, another important direction is system level communication control in FSL. Some studies reduce overall training cost by jointly considering model partitioning, communication resources, and computation resources. For example, Liang et al.~\cite{2026ITWC...25.1981L} propose a joint cutting point selection, communication allocation, and computation allocation strategy to balance convergence, latency, privacy, and resource constraints. However, this line of work mainly focuses on resource management and split point optimisation, while the frequent exchange of smashed activations and gradients can still dominate training cost when activations are high dimensional.

Another related direction is dynamic activation transmission control. SplitCom~\cite{li2026splitcomcommunicationefficientsplitfederated} exploits temporal redundancy in activations during split federated fine tuning of LLMs and uses adaptive threshold control to decide whether new activations should be uploaded or reused from cache. This reduces activation transmission frequency, but it focuses on LLM fine tuning rather than multi-node IoT time-series learning, and does not jointly adapt compression level and synchronisation interval.

Zhang et al.~\cite{10791300} propose a lightweight FedSL scheme with client side model pruning, gradient quantisation, activation dropout, and periodic aggregation. They derive a convergence upper bound characterising the joint effect of pruning rate, quantisation precision, aggregation interval, and split layer selection. Their results show that more frequent aggregation (smaller $I$) improves convergence, as pruning and quantisation act as implicit regularisers whose effect is best realised at $I=1$. However, the aggregation interval is fixed before training and does not adapt to runtime conditions, and the evaluation is mainly based on simulation.

Overall, existing studies either optimise resource allocation, reduce activation transmission frequency, or analyse fixed aggregation intervals. There is still limited work on joint runtime control of both activation compression and synchronisation interval in practical multi-node IoT FSL systems.

\subsection{Machine Learning for Rainfall Prediction in IoT Environments}

Machine learning has been widely used for rainfall prediction and environmental sensing with IoT data~\cite{saeed2021rainfall}. However, rainfall prediction remains difficult because of temporal variation and the skewed distribution of rainfall amounts, where heavy events are much rarer than light or moderate rain. Many studies focus on accuracy without considering communication cost, device heterogeneity, or distributed training constraints under FSL.
 
\section{Methodology}
\label{sec:Methodology}

\subsection{System Overview}
\label{subsec:overview}

The proposed FSL system comprises 11 edge clients and a central server communicating exclusively over gRPC. Four RPCs define the training lifecycle: \textit{Register} for client identification, \textit{Forward} for per step activation exchange, \textit{Synchronize} for FedAvg aggregation after every $\rho$ local epochs, and \textit{NotifyCompletion} for termination. In each \textit{Forward} call, the client sends a compressed smashed activation and reported latency; the server computes the loss, performs the backward pass on the server head, and returns the cut layer gradient with scheduler directives. Communication control resides on the server, so policy changes in the \textit{Forward} response propagate consistently without client side scheduling decisions.

\begin{figure*}[!t]
  \centering
  \includegraphics[width=0.75\linewidth]{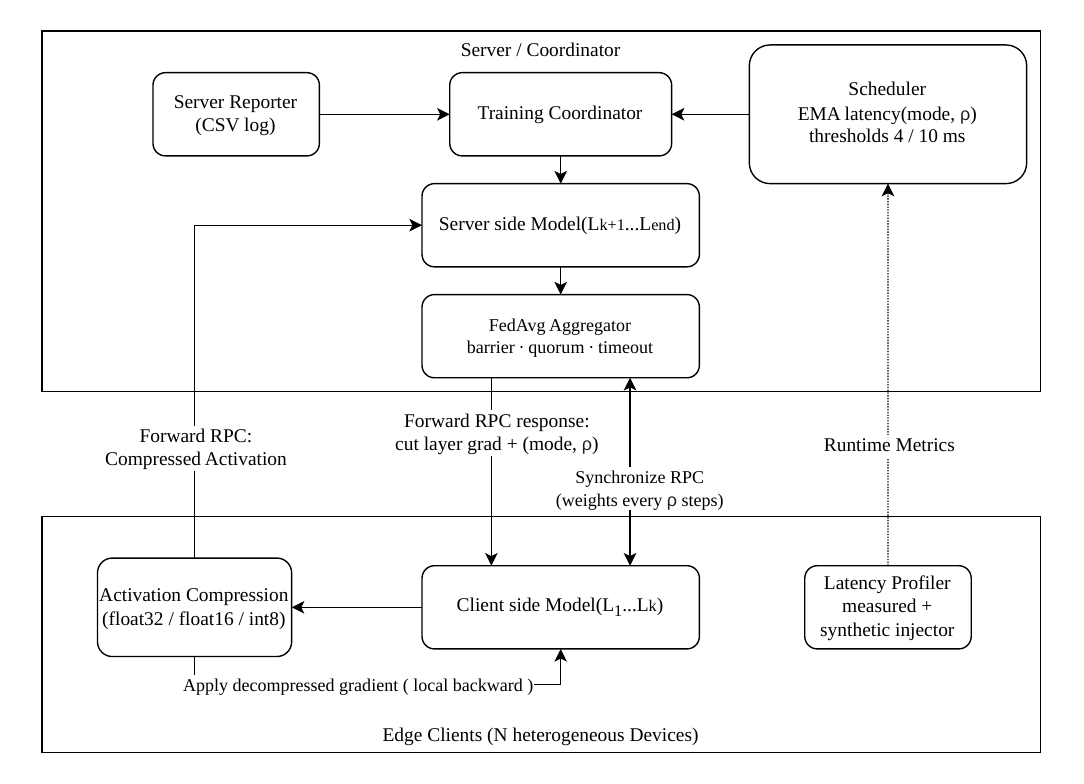}
  \caption{FSL system architecture: server coordinator with adaptive scheduler and server-side model; edge clients with encoder, compression module, and latency profiler.}
  \label{fig:architecture}
\end{figure*}

\subsection{Split Model Design}
\label{subsec:model}

The split model follows an encoder predictor decomposition: the client side module produces a fixed size smashed activation, and the server-side module completes prediction.

\subsubsection{Client side Encoder}

The client side \textit{ClientLSTM} is a two layer LSTM that maps each input window of shape $(\text{batch},\,48,\,5)$, representing 48 hourly steps across 5 meteorological features, to a fixed length smashed activation of shape $(\text{batch},\,64)$. The 64 dimensional output is a deliberate design choice: under float32 representation, each activation vector occupies exactly $64 \times 4 = 256$ bytes, making the per step activation upload payload fully deterministic.

\subsubsection{Server side Prediction Head}
The server-side \textit{ServerHead} is a two branch MLP that takes the 64 dimensional smashed activation and jointly produces a rain occurrence classifier and an auxiliary rainfall amount regression output. The auxiliary regression loss is applied only to samples with a positive rain label, regularising the shared representation around the long tail of large rainfall amounts (the regression target's skewed distribution, distinct from the near balanced occurrence label). This dual head design increases server-side computation but leaves the transmitted activation unchanged.

\subsection{Training Protocol}
\label{subsec:protocol}

The training protocol operates at two levels: the per step \textit{Forward} RPC flow in Section~\ref{subsec:overview} and a per epoch outer loop for global aggregation. One local epoch consists of 10 local optimisation steps, and $\rho$ controls the number of local epochs between synchronisation checks.

At the end of each local epoch, the client evaluates the condition
\begin{equation}
    (e + 1)\;\bmod\;\rho_{\text{current}} = 0,
    \label{eq:sync_trigger}
\end{equation}
where $e$ is the zero indexed epoch counter. If the condition is satisfied, the client issues a \textit{Synchronize} RPC with its local encoder weights, the global model round on which those weights were based, and the number of local epochs completed since the last refresh. The server maintains a timeout based aggregation barrier. In the reported experiments the nominal quorum is all 11 active clients, so static strategies normally behave as strict synchronous FedAvg. However, the implementation does not wait indefinitely: once a barrier window expires, the server aggregates the updates currently buffered for that round and immediately broadcasts the resulting global encoder. Clients that arrive after such a timeout receive the latest global weights; their update is accepted only if its base round is within the bounded staleness window, otherwise it is treated as a refresh only synchronisation. This matters for joint adaptive scenarios because different clients may operate under different $\rho$ values simultaneously, so submissions can reference an earlier server round. For each accepted update, aggregation follows staleness discounted local epoch weighted FedAvg:
\begin{equation}
    w_i = \frac{e_i}{1 + s_i},
    \label{eq:fedavg_weight}
\end{equation}
where $e_i$ is the client's number of completed local epochs and $s_i$ is its staleness; updates exceeding $S_{\max}$ are rejected and the client only refreshes to the latest global encoder. With $S_{\max}=0$ this reduces to standard local epoch weighted FedAvg, for which non-IID convergence under bounded gradient variance is established by Li et al.~\cite{Li2020FedAvg}; with $S_{\max}>0$ the protocol shares the bounded staleness setting analysed by Nguyen et al.~\cite{Nguyen2022FedBuff} for buffered asynchronous aggregation, we use $S_{\max}=3$ for adaptive scenarios to absorb up to three rounds of $\rho$-induced drift. The two scheduler outputs are applied at different granularities: the compression mode takes effect immediately at the next forward step, while the updated $\rho$ only affects the synchronisation check at the next epoch boundary, since altering the aggregation schedule mid epoch would disrupt the barrier consistency guarantee.

\subsection{Compression Modes}
\label{subsec:compression}

Three compression modes are applied symmetrically to both tensor directions of each \textit{Forward} RPC: \textbf{float32} (32 bit IEEE 754, 256~B per 64 dimensional activation), \textbf{float16} (16 bit half precision, 128~B), and \textbf{int8} (8 bit uniform quantisation with a 4 byte scale, 68~B). Symmetric application means the cut layer gradient returned from server to client uses the same mode as the activation upload. The payload metrics reported in this paper count the uploaded smashed activation. Since the returned cut layer gradients have the same shape and encoding, the bidirectional tensor payload of a forward and backward step is approximately twice the reported activation upload value, excluding RPC framing overhead.

\subsection{Adaptive Scheduler}
\label{subsec:scheduler}

The adaptive scheduler runs on the server and adjusts both the compression mode and the synchronisation interval $\rho$ at every forward step.

Each client's latency is tracked independently via an EMA:
\begin{equation}
    \hat{l}_t =
    \begin{cases}
        l_t & \text{if } \hat{l}_{t-1} = 0 \\[4pt]
        \alpha\,l_t + (1-\alpha)\,\hat{l}_{t-1} & \text{otherwise}
    \end{cases}
    \label{eq:ema}
\end{equation}
with smoothing factor $\alpha = 0.2$. The EMA is updated only when the reported latency $l_t > 0$, leaving the state unchanged for clients that report zero latency (e.g., in no profiler scenarios); the first valid observation seeds the estimate to avoid a cold start bias from a zero initialised EMA. Per client tracking ensures that heterogeneous network conditions do not mask individual stragglers. The smoothed latency is mapped to a compression mode via a three level rule:
\begin{equation}
    \text{mode} =
    \begin{cases}
        \text{float32} & \hat{l}_t \leq 4\,\text{ms} \\[2pt]
        \text{float16} & 4\,\text{ms} < \hat{l}_t \leq 10\,\text{ms} \\[2pt]
        \text{int8}    & \hat{l}_t > 10\,\text{ms}
    \end{cases}
    \label{eq:comp_policy}
\end{equation}
reflecting three operating regimes: full precision under low latency, half-precision activation upload payload halving when latency is moderate, and aggressive quantisation when latency is high.

The synchronisation interval is derived from the same severity index used for compression:
\begin{equation}
    \rho = \mathrm{clip}\!\left(
        \rho_{\text{base}} + \text{severity} \times \rho_{\text{step}},\;
        \rho_{\text{min}},\;
        \rho_{\text{max}}
    \right)
    \label{eq:rho_policy}
\end{equation}
where severity $\in \{0,1,2\}$ corresponds to the float32, float16, and int8 regimes; with $\rho_{\text{base}}=1$ and $\rho_{\text{step}}=1$, the three severity levels map to $\rho \in \{1, 2, 3\}$. The upper bound $\rho_{\text{max}}=20$ is a safety cap not reached in the current experiment matrix. Coupling $\rho$ to the same severity index as compression ensures that under high latency, both activation upload payload and synchronisation frequency are reduced simultaneously. The scheduler operates per step rather than per epoch so that it can respond to latency bursts that resolve within a single epoch; its $\mathcal{O}(1)$ overhead per step is negligible relative to the forward pass, and the rule based design is fully interpretable and deterministic.

\section{Experimental Setup}
\label{sec:experimental_setup}

\subsection{Dataset}
\label{subsec:dataset}

We evaluate the proposed FSL system using ERA5 based hourly observations from the Open Meteo Historical Weather API for Newcastle upon Tyne, UK. Data from 11 geographic weather stations are assigned one station per client, so each client stores only local meteorological observations. The input features are temperature, humidity, pressure, wind speed, and rain; the station wise partition creates a mildly heterogeneous IoT sensing scenario. Across eligible 48-hour window samples, the station level positive rate for the 24-hour rainfall label ranges from 49.46\% to 50.62\%, mean future 24-hour rainfall ranges from 2.23 to 2.36~mm, and the 95th percentile ranges from 10.10 to 10.60~mm. In the held out test period, station level positive rates range from 45.71\% to 46.83\%.

The dataset covers 2015-01-01 to 2026-03-31 at hourly resolution and is split chronologically by raw hourly rows: training before 2024-01-01 (78{,}888 rows per station, 80.0\%), validation over 2024 (8{,}784 rows, 8.9\%), and test from 2025-01-01 to 2026-03-31 (10{,}920 rows, 11.1\%). This chronological partition simulates real deployment conditions, training on historical observations and evaluating on strictly future unseen data to avoid temporal leakage.

Each sample uses the previous 48 hours of observations to predict rainfall over the next 24 hours. The binary label is positive if cumulative future rainfall is at least 0.5 mm, a threshold chosen to reduce sensitivity to trace drizzle. The regression target is the same future rainfall amount, trained in \texttt{log1p} space to reduce the influence of large values. Implementation, experiment configuration files, and data preparation scripts are available online\footnote{\url{https://gitfront.io/r/artifact-review-2026/hj4wQjs78x5q/csc8114/}}.

\subsection{System Configuration}
\label{subsec:system_configuration}

All scenarios use the same model architecture and training hyperparameters. The model is trained for up to 50 federated rounds at a learning rate of $5\times 10^{-4}$, with 10 local optimisation steps per client in each local epoch, and early stopping patience of 15 rounds.

The joint objective combines focal loss for rainfall occurrence classification and MSE loss for rainfall amount regression. The classification loss uses focal $\gamma=2.0$ and disables the focal $\alpha$ reweighting term. The classification and regression losses are weighted by 2.0 and 1.0, respectively. Since the occurrence label is close to balanced, the training sampler is not used to correct a severe class imbalance; instead, it fixes the rain positive sampling probability at 45\% to keep the occurrence and amount branches exposed to a consistent mixture of dry and rainy windows across clients and seeds.

The system contains 11 clients. The nominal minimum quorum is configured to 11 clients, with a 20~s barrier timeout and a 1~s grace period after the quorum is reached. Thus, static scenarios are effectively fully synchronous when every client reaches the barrier on time. However, the implementation can fall back to timeout based partial aggregation, followed by bounded staleness acceptance or refresh only synchronisation for late clients.

For simulation scenarios, network latency is emulated using a per client profiler that generates controlled inputs to the scheduler rather than measured network traces. Motivated by the broad latency variation observed in edge and IoT networks~\cite{8016573}, these inputs exercise the scheduler under three regimes: no injected latency, moderate latency near the float16/int8 decision boundary, and high latency that consistently selects the most aggressive policy. The no latency profile disables the profiler entirely. The low and high latency profiles use nominal base values of 8~ms and 50~ms with per client offsets spread up to $+6$~ms and $+30$~ms respectively (jitter std 1~ms and 5~ms); these values are stress test levels rather than claims about a specific wireless standard. The mixed latency profile assigns 4 clients to 0~ms, 4 to 8~ms, and 3 to 50~ms simultaneously (jitter std 2~ms). The Pi deployment disables the profiler and relies on the actual round trip delay between Pi clients and the cloud server.

The simulation experiments are conducted on a single Hetzner CPX52 cloud instance running Ubuntu 24.04.4 LTS, equipped with a 12-vCPU AMD EPYC-Genoa processor at 2.0~GHz and 22~GB of RAM, where all clients and the server execute as separate processes on the same host. The Pi deployment uses 11 Pi 4 Model B (quad-core ARM Cortex-A72, 4~GB RAM) devices as clients connected to a separate cloud server with 4~vCPUs and 8~GB of RAM over a real wide area link.The reduced server specification relative to the simulation host may contribute to longer per-step processing times on Pi. However, since all scenarios share the same Pi server, relative comparisons across P1--P4 remain valid. All processes run on CPU only, communicating through gRPC.

\subsection{Experiment Design}
\label{subsec:experiment_design}

The experimental design contains a controlled simulation matrix and a Pi deployment matrix. Simulation isolates compression and synchronisation effects under matched injected latency profiles, while the Pi deployment tests the same strategies on physical edge devices over a real client--server network. The simulation matrix varies latency profile (no, low, high, and mixed latency) and communication strategy (static compression, static synchronisation interval control, adaptive compression, and joint adaptive control).

\begin{table}[!tbp]
\centering
\caption{Simulation scenario matrix. Strategies S0--S3 use fixed configurations with the scheduler disabled; S4 enables compression only adaptation; S5 enables joint compression and $\rho$ adaptation.}
\label{tab:native_simulation_matrix}
\resizebox{\columnwidth}{!}{%
\small
\begin{tabular}{lcccc}
\toprule
\textbf{Strategy} & \textbf{No Latency} & \textbf{Low ($\sim$8 ms)} & \textbf{High ($\sim$50 ms)} & \textbf{Mixed} \\
\midrule
S0: float32, $\rho{=}1$ (baseline) & N01 & L05 & H11 & --  \\
S1: float16, $\rho{=}1$            & N02 & L06 & H12 & --  \\
S2: int8, $\rho{=}1$               & N03 & L07 & H13 & --  \\
S3: float32, $\rho{=}3$            & N04 & L08 & H14 & --  \\
S4: adaptive compression, $\rho{=}1$ & --  & L09 & H15 & --  \\
S5: joint adaptive                 & --  & L10 & H16 & M17 \\
\bottomrule
\end{tabular}%
}
\end{table}

Strategies are compared under the same latency profile wherever possible, while no latency scenarios provide clean ablations for float16, int8, and $\rho{=}3$. Scenario M17 applies joint adaptation to a mixed profile with 4 clients at no latency, 4 at $\sim$8~ms, and 3 at $\sim$50~ms, testing whether the scheduler responds to heterogeneous client conditions within the same round.

The Pi deployment uses the same FSL software stack but runs clients on physical Pi devices and the server on a cloud VPS. Unlike the simulation, no latency profiler is used; the measured baseline round trip time between clients and the server is approximately 21--24~ms. The scheduler thresholds of 4~ms and 10~ms are intentionally kept identical to the simulation rather than retuned for this baseline. Because the measured RTT consistently exceeds 10~ms, this configuration places the Pi deployment in the most aggressive endpoint of the adaptive policy (int8 compression with $\rho{=}3$) for the entirety of training. The intention is therefore not to demonstrate frequent switching on Pi, but to stress test the high latency policy endpoint selected by the same scheduler. If the most aggressive scheduled state does not degrade predictive quality under real wide area conditions, this suggests that intermediate scheduler states observed in simulation are unlikely to be more harmful in this setting.

Each scenario is repeated with three random seeds: 42, 52, and 62. These seeds are chosen with fixed offsets to provide independent initialisations and avoid bias from a single seed. Evaluating across these seeds helps reduce random variations from model initialisation and data shuffling. The main simulation matrix contains 17 scenarios, while the Pi matrix contains four representative scenarios mapped to the main strategy groups: P1 to the float32 baseline (S0), P2 to float16 compression (S1), P3 to fixed $\rho{=}3$ synchronisation control (S3), and P4 to adaptive joint control (S5).

\subsection{Evaluation Metrics}
\label{subsec:evaluation_metrics}

We report prediction quality, communication efficiency, and system efficiency to analyse the trade-off between predictive performance and system cost.

For prediction quality, the primary classification metric is AUPRC. AUPRC is threshold free and is more suitable than a single threshold F1 score when comparing scenarios where different compression modes may shift the model output distribution. We also report ROC-AUC as a secondary threshold free metric and F1 score at the configured probability threshold of 0.5 for completeness. The rainfall amount branch is used only as auxiliary regularisation for the shared representation, and no deployment decision in this study uses the regression output as a standalone target; regression error is therefore not reported. Evaluation focuses on occurrence prediction and communication system trade-offs.

All F1 values use a fixed probability threshold of 0.5 across every scenario and seed. We avoid per scenario threshold tuning because it would entangle communication strategy effects with threshold optimisation; threshold free AUPRC and ROC-AUC are therefore treated as the primary indicators of predictive quality.

For communication efficiency, we measure activation upload payload, synchronisation traffic for periodic FedAvg updates, and effective per client data throughput. Simulation communication totals are reported per client, while Pi communication totals are aggregated across all 11 clients and accompanied by per client interpretations in the text.

For system efficiency, we report samples processed per second, end-to-end runtime, and average per step round trip latency. For the Pi deployment, peak memory usage is also recorded to assess feasibility on constrained edge hardware.

\section{Results and Analysis}
\label{sec:Res}

\subsection{Simulation Results}
\label{subsec:sim_results}

The simulation matrix evaluates 17 scenarios across three injected latency profiles (no latency, low $\sim$8~ms, and high $\sim$50~ms) and six communication strategies. Metrics are computed per client on 500 held out test samples; AUPRC is reported as mean $\pm$ std across 3 independent seeds, and other metrics are seed means.

\begin{figure}[!tbp]
  \centering
  \includegraphics[width=\linewidth]{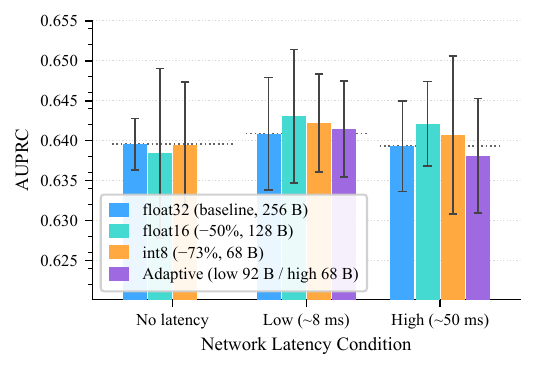}
  \caption{AUPRC by compression mode and latency condition (mean $\pm$std, 3 seeds). Dotted lines mark the float32 baseline.}
  \label{fig:compression_auprc}
\end{figure}

AUPRC is stable across all 17 scenarios (0.6381--0.6484), indicating that neither compression mode nor synchronisation interval materially degrades predictive quality. The spread (0.010) is narrower than the seed level standard deviation of several scenarios (e.g., N02: $\pm$0.011). With only three seeds and observed AUPRC std in the range $\pm$0.0001--$\pm$0.011, the minimum detectable effect at $\alpha{=}0.05$ ($df{=}2$) is on the order of $0.02$--$0.04$, which is well above the observed across scenario spread. We therefore, do not perform per pair significance testing, and treat AUPRC differences, including the numerically higher $\rho{=}3$ scenarios (N04, L08, H14), as suggestive of robustness rather than confirmed ranking. Joint adaptive scenarios (L10, H16) remain within seed variance of their $\rho{=}3$ counterparts while additionally reducing activation upload payload, as Fig.~\ref{fig:compression_auprc} confirms.

\begin{table*}[!tbp]
\centering
\caption{Simulation results: prediction quality and per client communication cost (AUPRC: mean $\pm$ std across 3 seeds; other metrics: seed means).}
\label{tab:sim_results}
\small
\begin{tabularx}{\linewidth}{lXcccrrr}
\toprule
    & & \multicolumn{3}{c}{\textbf{Prediction Quality}} & \multicolumn{3}{c}{\textbf{Communication Cost}} \\
\cmidrule(lr){3-5}\cmidrule(lr){6-8}
\textbf{Scenario} & \textbf{Strategy} & \textbf{AUPRC} & \textbf{ROC-AUC} & \textbf{F1} & \makecell[r]{\textbf{Activation Upload}\\\textbf{(B/step)}} & \makecell[r]{\textbf{Activation Upload}\\\textbf{(MB/client)}} & \makecell[r]{\textbf{Sync}\\\textbf{(MB/client)}} \\
\midrule
\multicolumn{8}{l}{\textit{No latency}} \\
N01 & float32, $\rho{=}1$ (baseline)   & 0.6396 $\pm$ 0.0032 & 0.7101 & 0.6023 & 256.0 & 4.21 & 6.74 \\
N02 & float16, $\rho{=}1$              & 0.6384 $\pm$ 0.0106 & 0.7098 & 0.6052 & 128.0 & 2.19 & 7.02 \\
\textbf{N03} & \textbf{int8, $\rho{=}1$} & 0.6395 $\pm$ 0.0078 & 0.7104 & 0.5688 & \textbf{68.0} & \textbf{1.15} & 6.95 \\
\textbf{N04} & \textbf{float32, $\rho{=}3$} & \textbf{0.6473} $\pm$ 0.0001 & \textbf{0.7166} & 0.6075 & 256.0 & 2.08 & \textbf{3.19} \\
\midrule
\multicolumn{8}{l}{\textit{Low latency ($\sim$8 ms)}} \\
L05 & float32, $\rho{=}1$ (baseline)   & 0.6409 $\pm$ 0.0070 & 0.7124 & 0.5293 & 256.0 & 4.49 & 7.19 \\
L06 & float16, $\rho{=}1$              & 0.6431 $\pm$ 0.0083 & 0.7155 & 0.5831 & 128.0 & 2.16 & 6.92 \\
L07 & int8, $\rho{=}1$                 & 0.6422 $\pm$ 0.0061 & 0.7137 & 0.6129 &  68.0 & 1.15 & 6.90 \\
L08 & float32, $\rho{=}3$              & 0.6479 $\pm$ 0.0004 & 0.7191 & 0.6185 & 256.0 & 2.08 & 3.19 \\
L09 & adaptive compression           & 0.6415 $\pm$ 0.0060 & 0.7131 & \textbf{0.6358} &  92.5 & 1.55 & 6.86 \\
\textbf{L10} & \textbf{joint adaptive} & \textbf{0.6474} $\pm$ 0.0004 & \textbf{0.7163} & 0.5815 & \textbf{92.5} & \textbf{0.92} & \textbf{3.79} \\
\midrule
\multicolumn{8}{l}{\textit{High latency ($\sim$50 ms)}} \\
H11 & float32, $\rho{=}1$ (baseline)   & 0.6393 $\pm$ 0.0057 & 0.7106 & 0.6085 & 256.0 & 4.24 & 6.79 \\
H12 & float16, $\rho{=}1$              & 0.6421 $\pm$ 0.0053 & 0.7131 & \textbf{0.6308} & 128.0 & 2.16 & 6.91 \\
H13 & int8, $\rho{=}1$                 & 0.6407 $\pm$ 0.0099 & 0.7145 & 0.5560 &  68.0 & 1.12 & 6.75 \\
H14 & float32, $\rho{=}3$              & 0.6484 $\pm$ 0.0004 & 0.7194 & 0.6265 & 256.0 & 2.08 & 3.19 \\
H15 & adaptive compression           & 0.6381 $\pm$ 0.0072 & 0.7098 & 0.5640 &  68.0 & 1.11 & 6.66 \\
\textbf{H16} & \textbf{joint adaptive} & \textbf{0.6483} $\pm$ 0.0003 & \textbf{0.7195} & 0.6222 & \textbf{68.0} & \textbf{0.55} & \textbf{3.19} \\
\midrule
\multicolumn{8}{l}{\textit{Mixed latency}} \\
M17 & joint adaptive (mixed)         & 0.6476 $\pm$ 0.0007 & 0.7179 & 0.6368 & 128.2 & 1.45 & 4.36 \\
\bottomrule
\end{tabularx}
\end{table*}

On communication, int8 gives the lowest per step activation upload payload (68~B), $\rho{=}3$ reduces per client synchronisation traffic by $\sim$53\% ($\sim$6.7~MB to 3.19~MB), and joint adaptive combines both: H16 reaches the lowest per client activation upload payload (0.55 MB) and sync (3.19~MB), while adaptive compression only scenarios reduce activation upload payload but not sync traffic. Simulation wall clock runtime is not treated as primary evidence because all clients and the server share one cloud host, so process scheduling and CPU contention dominate the measured time more than realistic network transfer. The simulation runtime values in Table~\ref{tab:pi_combined} are therefore used only as cross platform context; the Pi deployment provides the runtime evidence.

\begin{figure}[!tbp]
  \centering
  \includegraphics[width=\linewidth]{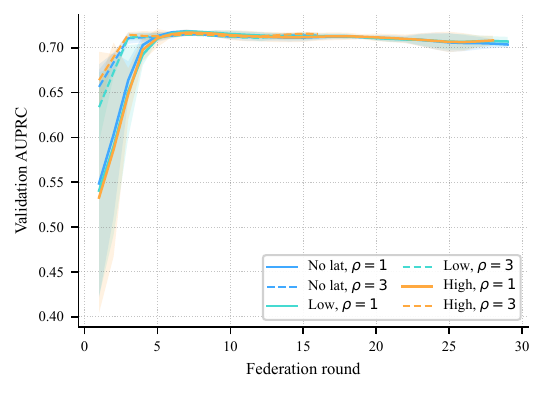}
  \caption{Validation AUPRC per federation round for $\rho{=}1$ (solid) and $\rho{=}3$ (dashed) across the three latency profiles (colour), $\pm$1 std across 3 seeds.}
  \label{fig:rho_convergence}
\end{figure}

Fig.~\ref{fig:rho_convergence} reports validation AUPRC per federation round, not per local optimisation step or wall clock second. Since a $\rho{=}3$ round contains three local epochs (each 10 steps) before aggregation, the earlier plateau of $\rho{=}3$ in federation round units reflects reduced synchronisation overhead under round based early stopping, where each barrier represents more local work, rather than improved optimisation efficiency per local step.

\begin{figure}[!tbp]
  \centering
  \includegraphics[width=\linewidth]{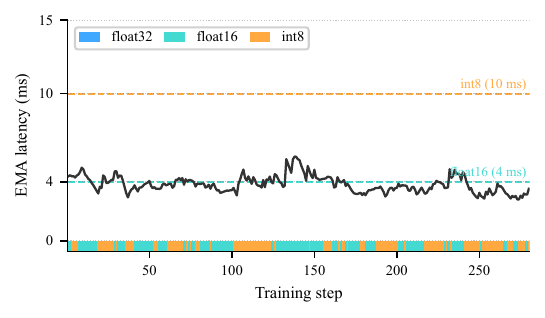}
  \caption{Per step EMA latency and assigned compression mode for Client~5 in scenario L09 (seed~42). Dashed lines mark the float16 (4~ms) and int8 (10~ms) thresholds.}
  \label{fig:scheduler_timeline}
\end{figure}

Fig.~\ref{fig:scheduler_timeline} illustrates the scheduler's real time behaviour for Client~5 in scenario L09 (seed~42), whose latency oscillates near the float16 and int8 boundary. The EMA smoothed trace closely tracks the threshold crossings and the assigned mode shifts accordingly, confirming that the implementation follows the intended threshold policy.

\subsection{Pi Deployment Results}
\label{subsec:pi_results}

Four representative scenarios (P1--P4) were deployed on physical Pi 4 B clients connected to a cloud server via real wide area links ($\sim$21--24~ms baseline RTT). Unlike the simulation, latency is not injected by a profiler but arises from actual client--server communication. As described in Section~\ref{subsec:experiment_design}, the unchanged scheduler thresholds (4~/~10~ms) place P4 in the int8\,+\,$\rho{=}3$ regime throughout training: empirical logs confirm that clients consistently maintain $\rho{=}3$ and int8 compression.

\begin{table*}[!tbp]
\centering
\caption{Pi deployment results with simulation cross reference (AUPRC and Pi runtime: mean $\pm$ std across 3 seeds; remaining metrics: seed means; communication columns are all client totals).}
\label{tab:pi_combined}
\small
\addtolength{\tabcolsep}{-2pt}
\begin{tabularx}{\linewidth}{Xccccccc ccc} 
\toprule
& \multicolumn{2}{c}{\textbf{Prediction Quality}} & \multicolumn{4}{c}{\textbf{Communication}} & \multicolumn{4}{c}{\textbf{System Efficiency}} \\
\cmidrule(lr){2-3}\cmidrule(lr){4-7}\cmidrule(lr){8-11}
\textbf{Scenario} & \makecell{\textbf{AUPRC}\\\textbf{(Pi)}} & \makecell{\textbf{AUPRC}\\\textbf{(Sim)}} & \makecell{\textbf{Activation Upload}\\\textbf{Total (MB)}} & \makecell{\textbf{Sync}\\\textbf{Total (MB)}} & \makecell{\textbf{Data Tput}\\\textbf{(kbps)}} & \makecell{\textbf{Reduc.}} & \makecell{\textbf{Tput}\\\textbf{SPS}} & \makecell{\textbf{Runtime}\\\textbf{Pi (s)}} & \makecell{\textbf{Sim}\\\textbf{(s)}} & \makecell{\textbf{Lat}\\\textbf{(ms)}} \\
\midrule
P1: float32, $\rho{=}1$ & 0.6381$\pm$.0052 & 0.6396$\pm$.0032 & 47.36 & 75.77 & 14.77 & -- & 1.72 & 3{,}280$\pm$688 & 609 & 26.93 \\
P2: float16, $\rho{=}1$ & 0.6434$\pm$.0053 & 0.6384$\pm$.0106 & 23.93 & 76.57 & 7.38 & $-$49\% & 1.72 & 3{,}318$\pm$814 & 624 & 26.89 \\
P3: float32, $\rho{=}3$ & 0.6479$\pm$.0008 & 0.6473$\pm$.0001 & 22.83 & 35.06 & 7.02 & $-$52\% & 1.65 & 3{,}331$\pm$26 & 499 & 26.94 \\
\textbf{P4: Adaptive} & \textbf{0.6482$\pm$.0014} & \textbf{0.6483$\pm$.0003} & \textbf{6.07} & \textbf{35.06} & \textbf{1.87} & \textbf{$-$87\%} & \textbf{1.66} & 3{,}317$\pm$10 & \textbf{504} & 27.27 \\
\bottomrule
\end{tabularx}
\end{table*}

All Pi scenarios maintain comparable AUPRC (0.638--0.648). P4 is numerically highest in AUPRC (0.648), but differences across all four scenarios remain within 0.011. The primary finding is therefore robustness under communication reduction rather than a statistically significant ranking difference.

On communication efficiency, P4 reduces the \textbf{total activation upload payload to 6.07~MB} (0.55~MB per client, $-$87\% vs.~P1) and the \textbf{total synchronisation traffic to 35.06~MB} (3.19~MB per client, $-$54\%), combining the strengths of compression only P2 ($-$49\% activation upload payload) and sparse synchronisation P3 ($-$54\% sync). P2's synchronisation traffic (76.57~MB) is essentially unchanged from P1 (75.77~MB), since both use $\rho{=}1$ and the 1\% delta sits within the $\sim$17~MB seed level std.

On system efficiency, mean throughput and runtime are comparable (1.65--1.72~SPS; 3,280--3,331~s), as reduced synchronisation overhead in P3/P4 is balanced by adaptation and compression cost. The key differentiator is \textbf{runtime stability}: P3 and P4 achieve runtime standard deviations of $\pm$26~s and $\pm$10~s versus $\pm$688~s and $\pm$814~s for P1 and P2.

\subsection{Simulation vs.\ Pi: Cross platform Comparison}
\label{subsec:comparison}

\begin{figure}[!tbp]
  \centering
  \includegraphics[width=\linewidth]{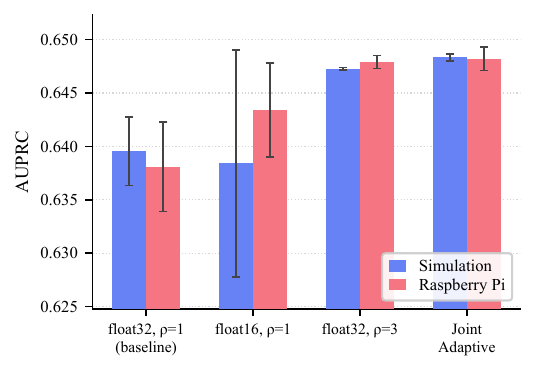}
  \caption{Simulation vs.\ Pi AUPRC for four communication strategies ($\pm$1 std, 3 seeds).}
  \label{fig:cross_platform}
\end{figure}

Table~\ref{tab:pi_combined} pairs each Pi scenario with its closest simulation counterpart. Pi latency ($\sim$27~ms) falls between the low ($\sim$8~ms) and high ($\sim$50~ms) simulation profiles; P4 is paired with H16 because both converge to the same int8$+\rho$=3 operating point. Fig.~\ref{fig:cross_platform} visualises the AUPRC comparison.

AUPRC differences between Pi and simulation remain within 0.007. The $\rho{=}3$ and joint adaptive scenarios are numerically higher than the $\rho{=}1$ baselines on both platforms, but these small gaps are treated as robustness evidence rather than statistically significant ranking differences. Joint adaptive control nevertheless achieves the best combined activation upload and synchronisation reduction on both platforms. The platform dependent difference is runtime: Pi mean runtimes are about 5$\times$ longer than simulation (3,280~s vs 609~s for the float32 baseline), and $\rho{=}3$ or joint adaptive do not reduce Pi mean runtime because reduced synchronisation overhead is offset by longer local computation.

\section{Conclusions}
\label{sec:Conclusion}

We presented an FSL framework that jointly regulates activation compression and the synchronisation interval $\rho$ using a lightweight EMA-driven server-side scheduler, evaluated through a 17 scenario simulation matrix and a four scenario Pi deployment. The simulation scenarios exercise runtime scheduler switching, while the Pi deployment validates the high latency endpoint selected by the same policy on physical devices.

The empirical results support three conclusions. First, AUPRC is stable across compression modes and synchronisation intervals (0.6381--0.6484 in simulation; within 0.011 on Pi), indicating that int8 quantisation and $\rho{=}3$ do not materially degrade predictive quality in this setting. Second, joint adaptive control provides the best combined communication reduction in simulation: H16 reaches the lowest activation upload payload (0.55~MB) and sync traffic (3.19~MB), while the Pi high latency endpoint selected by the policy achieves 87\% activation upload and 54\% synchronisation reductions. Third, less frequent synchronisation reaches the early stopping criterion after fewer aggregation barriers and substantially stabilises real wide area runtime ($\pm$10~s for P4 versus $\pm$688~s for P1), even when mean runtime is unchanged.

Several limitations remain. The rule based scheduler could be replaced by a learning based or control theoretic policy using signals such as bandwidth and queue depth. Although the implementation supports timeout based partial aggregation, the evaluation does not intentionally model client dropout, so more systematic dropout tolerant experiments are required before deployment. Larger deployments and additional IoT time-series tasks are also needed to test generalisation, while dynamic cut layer selection is left for future work.


\bibliographystyle{IEEEtran}
\bibliography{IEEEabrv,refs}

\end{document}